\pdfoutput=1

\documentclass[11pt]{article}

\usepackage{acl}

\usepackage{times}
\usepackage{latexsym}

\usepackage[T1]{fontenc}

\usepackage[utf8]{inputenc}

\usepackage{microtype}

\usepackage{inconsolata}

\usepackage{graphicx}

\title{When are We Worried?\\ Temporal Trends of Anxiety and What They Reveal about Us}

\author{Saif M. Mohammad \\
  National Research Council Canada \\
  \texttt{saif.mohammad@nrc-cnrc.gc.ca} \\}

\begin{document}
\maketitle
\begin{abstract}
In this short paper,
we make use of a recently created lexicon of word--anxiety associations
to analyze large amounts of US and Canadian social media data (tweets) to explore
\textit{*when*} we are anxious and what insights that reveals about us.
We show that our levels of anxiety on social media exhibit systematic
patterns of rise and fall during the day --- highest at 8am (in-line with when we have high cortisol levels in the body) and lowest around noon. 
Anxiety is lowest on weekends and highest mid-week.
We also examine anxiety in past, present, and future tense sentences to show
that anxiety is highest in past tense and lowest in future tense.
Finally, we examine the use of anxiety and calmness words in posts that contain pronouns to show: 
more anxiety in 3rd person pronouns (\textit{he, they}) posts than 1st and 2nd person pronouns
and higher anxiety in posts with subject pronouns (\textit{I, he, she, they}) than object pronouns (\textit{me, him, her, them}). 
Overall, these trends provide valuable insights on not just when we are anxious, but also how different types of focus (future, past, self, outward, etc.) are related to anxiety.
\end{abstract}


\section{Introduction}

Anxiety is a healthy human emotion that has considerable impact on our 
 safety, well-being, health, behaviour, and mood.
It is the apprehensive uneasiness or nervousness due to some anticipated or potential negative outcome. 
Psychologists have long posited its evolutionary benefits \cite{bateson2011anxiety};
yet, there is much we do not know. 
Further, some have argued that 
the dramatic technological advances over the last few decades have rapidly and substantially changed our environments to such an extent (for instance, the widespread social media usage and human--technology interactions) that there is now a considerable mismatch between the current environment and what our anxiety response slowly evolved to address over millennia \cite{blease2015too,salecl2004anxiety,qureshi2022anxiety}. 

Interestingly, the widespread use also means that our language on social media can be a powerful window into our anxiety. In this short paper, we focus on social media posts to examine \textit{*when*} we are anxious and what insights that reveals about us.
For this, we make use of a recently created repository of word--anxiety associations (WorryWords) \cite{mohammad-2024-worrywords} to analyze large amounts of posts on Twitter (now X).

\section{Related Work}

Over two decades of research exists on tracking emotions in social media 
for public health. Work on anxiety has been largely limited to anxiety disorders and depression (as opposed to general anxiety): e.g., the 
relationship between the language on social media and symptoms of anxiety \cite{o2021relationship,vedula2017emotional}.
Perhaps the work closest to this is the highly influential work of \newcite{dodds2011temporal}, who analyzed temporal patterns of hedonism/sentiment in tweets using a lexicon of word--happiness associations.
We use similar techniques but also bring in theories on focus (self focus, social focus, past focus, etc.) from psychology (discussed ahead) to reveal what temporal trends of anxiety reveal about us.

 \begin{figure*}[t!]
	\centering
	    \includegraphics[width=\textwidth]{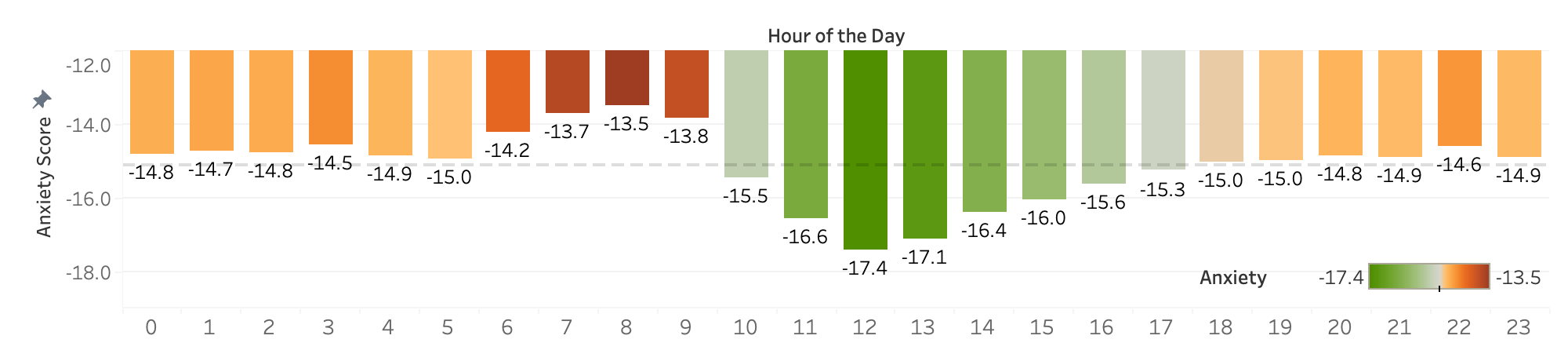}
        \caption{Aggregate anxiety scores of posts by hour of the day.}
	    \label{fig:anx-hour}
        \vspace*{-3mm}
\end{figure*}

\section{Temporal Trends of Anxiety}


Knowing \textit{*when*} we are more anxious is important because then we can better understand our anxiety and appropriate remedial actions can be taken (at the right time).
Psychology research, through self-reported data, has consistently shown that anxiety varies over time, especially throughout the day and across the week.
Notably, people are believed to be more anxious immediately after waking up (because of elevated cortisol levels and thinking about the work day), calmer in the afternoons (as they are entrenched in their routines), and again more anxious in the evenings (as they reflect on unfinished tasks, mistakes, unresolved issues, etc.) \cite{moberly2008ruminative,takano2013ruminative,cox2022effects,vedhara2003investigation}.
There are more mixed results on day of the week: weekends are believed to be periods of low anxiety but people are also believed to be anxious Sunday evenings as the work week approaches.
Some work points to anxiety being highest on Mondays and falling mid-week \cite{kao2014weekly,abu2014does}, whereas others say anxiety peaks mid-week  \cite{curran1997mood}.
Self-reports are also limited by biases (e.g., recency bias, social desirability bias, and temporal aggregation bias) \cite{caputo2017social}. 

Thus, looking at language in social media posts is a valuable complement to self-reported information. Even though what we speak or post may not always reflect how we feel, analyzing trends in large amounts of language data can provide useful insights: at minimum on trends in anxiety in our social media posts, and arguably more broadly on when we are likely to feel more or less anxious.

We use WorryWords (a large lexicon of word--anxiety associations) and TUSC (a corpus of tweets) \cite{vishnubhotla-mohammad-2022-tusc} in the ABCDE \cite{wahle2025abcde}    
 suite of datasets to explore when North Americans use more anxiety words. 
WorryWords includes entries for about 44,500 English words: about 26\% of these are marked as associated with anxiety and about 13\% are marked as associated with calmness. The authors report high annotation reliability (annotation--reannotation Pearson's correlations of 0.89).
TUSC includes lemmatized and lower-cased American and Canadian geo-located posts on X (formerly Twitter) from 2015 to 2021 \cite{vishnubhotla-mohammad-2022-tusc}. 

\subsection{Time and Anxiety: Existential Focus, Task Focus, Social Focus, Self-Evaluation Focus}
We conducted a simple experiment to examine the use of anxiety and calmness associated words in tweets posted at different hours of the day and in tweets posted on different days of the week. Specifically we mapped the time stamps of tweets in TUSC to local time and week day. All the tweets pertaining to different hours of the day (from 0 to 23) are put in corresponding (24) bins. All the tweets pertaining to different days of the week 
are put in corresponding (7) bins.

 For each bin, we obtained aggregate-level anxiety scores by examining the WorryWords entries of terms in the posts.
 Following \cite{teodorescu-mohammad-2023-evaluating,turney-2002-thumbs} we calculated the percentage of anxiety-associated words minus the percentage of calmness-associated words.\footnote{\citet{teodorescu-mohammad-2023-evaluating} and \citet{turney-2002-thumbs} show that this formula accurately captures temporal trends of emotions in various text streams for valence, arousal, anger, sadness, etc. (Correlations of over 0.9 with gold emotion arcs.) Other similar formulae may also be be used. Our goal was to use a simple and interpretable approach that has been shown to work well for aggregate-level analysis.} 
 \textbf{The average anxiety score for the full dataset is: $-$15.13} (shown as a dashed line in the figures ahead).
 Note that this and other scores shown ahead are all negative indicating that on average people use more calmness-associated words than anxiety-associated words. 
For all the experiments below, we used t-test to test for statistically significant differences, using a significance level of 0.05.

 \subsubsection{Hour of Day and Anxiety}

Figure \ref{fig:anx-hour} shows the aggregated anxiety scores for tweets posted in each of the hours from 0 to 23.   Observe that tweets with the most anxiety-associated and fewest calmness associated words are posted in the morning (anxiety score peaking to $-$13.5 at 8am).
In contrast, we post tweets with the fewest anxiety-associated and most calmness associated words around noon (anxiety score dipping to $-$17.4); after which, the anxiety score steadily increases to close-to-peak levels at 10pm.
The difference between various pairs of scores was found to be statistically significant, including: 5--8am, 8--10am, 10am--12pm, 12--2pm, and 2--6pm.

Overall, these results on tweets are highly consistent with research on self-reported anxiety levels. Thus, they add to the evidence on how our biology (cortisol levels), work focus, and existential focus impact anxiety at  varies times of the day. They also show for the first time that, at an aggregate level, the anxiety conveyed in our posts on social media reflect our levels of anxiety.

\subsubsection{Day of the Week and Anxiety}

\begin{figure}[t]
	\centering
	    \includegraphics[width=0.4\textwidth]{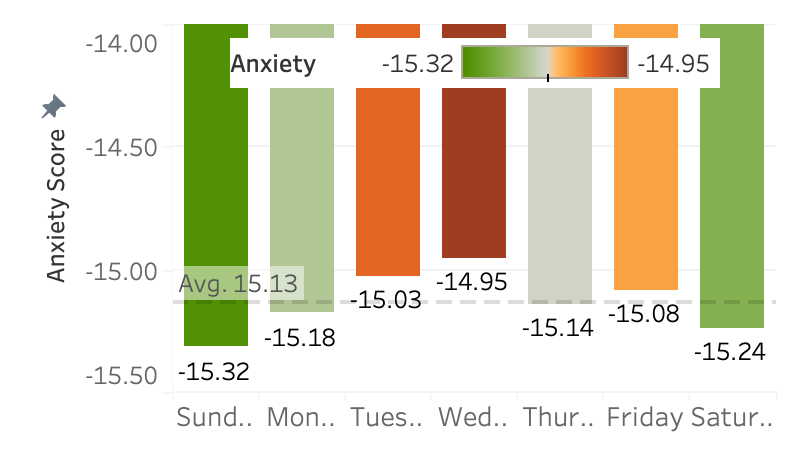}
       \vspace*{-3mm}
        \caption{Anxiety scores of posts by day of week.}
	    \label{fig:anx-dayweek}
        \vspace*{-3mm}
\end{figure}

Figure \ref{fig:anx-dayweek} shows the aggregated anxiety scores for tweets posted in each of the days of the week. Observe that 
the  scores are lowest (indicating most calmness) on the weekend and highest (indicating most anxiousness) on Wednesday (mid week). The difference in scores between Sunday and Wednesday is statistically significant.

Thus, we do not see evidence of Sunday scaries\footnote{Creeping sense of anxiety towards the end of Sunday about the coming work week.} in these results; however, this may be because while there is some amount of anxiety due to the approaching work week, it is far outweighed by the calmness of the holiday spread throughout the day.
The peak levels of anxiety on Wednesday are consistent with the belief that people are more anxious when the rest of the past weekend has worn off and the coming weekend is still a few days away.

Note that the changes in anxiety score are much greater across the hours of the day ($-$17.4 to $-$13.5) than across the days of the week ($-$15.32 to $-$14.95). This indicates that the time of the day has a stronger impact on our anxiety levels than the day of the week.

\subsection{Past, Present, and Future Focus}

As noted earlier, anxiety is often described as an unease of something that may happen in the future. Several studies have also shown that anxiety is associated with a future focus \cite{miloyan2016episodic}. Other work has shown that anxiety is associated with present and past tenses---repetitive, rumination common in cases of social anxiety (e.g., \textit{my heart is racing; I feel they are judging me)} or recalling past failures and traumas that make one fear the future \cite{aastrom2019getting}. 

To shed more light on this, we wanted to explore the extent to which people use anxiety- and calmness associated words in tweets pertaining to different tenses. For this, we partition the TUSC posts into: tweets that have a past tense verb (deemed to be past tense tweets), tweets that have a future tense indicating modal or signal word (\textit{will, won't, shall, expect, believe, hope, tomorrow, next day/week/month/year}) and a present tense verb (deemed to be future tense tweets), and tweets that have a  present tense verb but no future tense modal/signal word (deemed to be present tense tweets).

\begin{figure}[t]
	\centering
	    \includegraphics[width=0.47\textwidth]{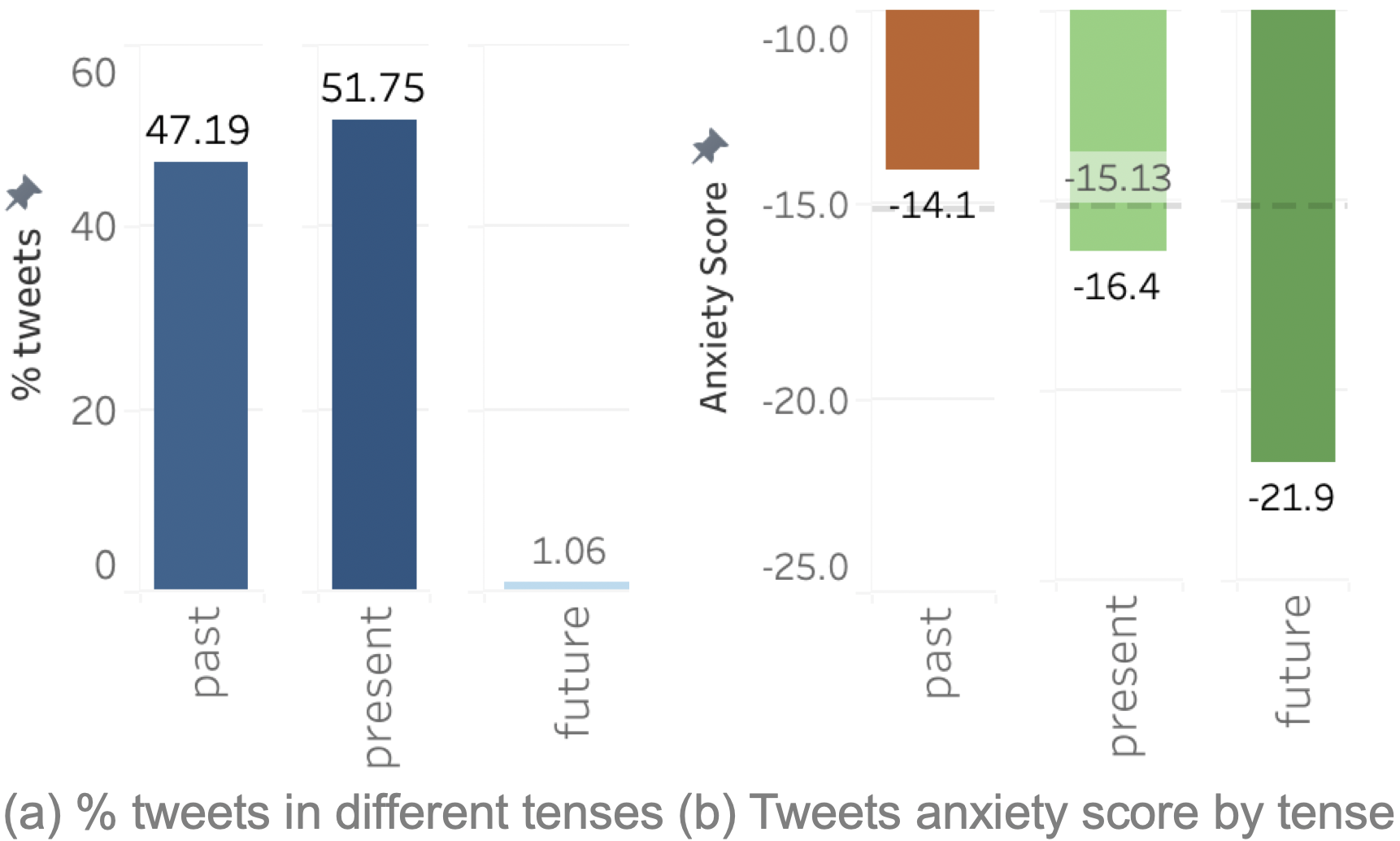}
        \caption{Tense and anxiety in tweets.}
	    \label{fig:anx-tense}
        \vspace*{-3mm}
\end{figure}

Figure \ref{fig:anx-tense} (a) shows the percentage of tweets pertaining to each of the three tenses. Observe that the most frequent tense in English tweets is the present tense ($\sim$52\%), closely followed by past tense   ($\sim$47\%), while only $\sim$1\% of the tweets are in future tense.

\begin{figure*}[t]
	\centering
	    \includegraphics[width=\textwidth]{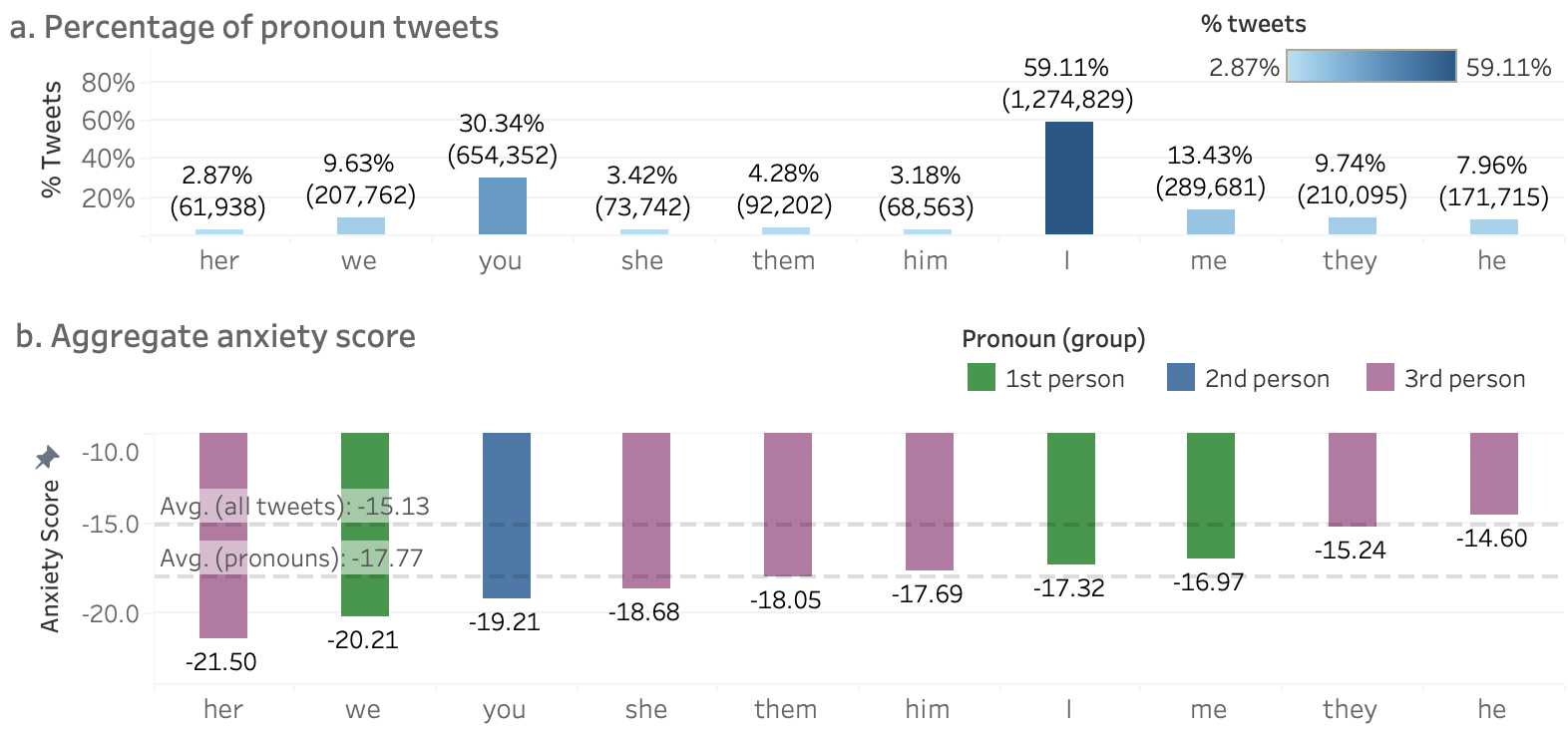}
        \vspace*{-3mm}
        \caption{Distribution of pronouns in tweets (a) and anxiety scores of  pronoun-including tweets (b).}
	    \label{fig:anx-pron}
\end{figure*}

Figure \ref{fig:anx-tense} (b) shows the aggregate anxiety scores for tweets in each of the tenses. We see that past tense tweets have the highest anxiety score ($-$14.1) whereas present tense ($-$16.4) and future tense ($-$21.9) have anxiety scores much lower than the average ($-$15.13). 
The differences in scores between all pairs of tenses was found to be statistically significant.

Overall these results show that anxiety is most often expressed in past tense. Even though anxiety has an inherent connection with future unease, it is expressed through utterances worrying about the past and present. Future tense tweets include substantially more calmness terms than anxiety terms. This is likely because of a wide-spread general optimism about the future (even if this optimism has decreased some what in recent years).

\subsection{Pronouns and Anxiety: Self Focus, Group/Relational Focus, Social Focus}

Pronoun use has emerged as a particularly useful indictor of one's mental health, emotions, and well-being. Notably, they reflect self-focus, social connection, and emotional perspective \cite{zimmermann2013way}. Emotional perspective refers to the mental stance one takes toward an experience. It includes: point of view (first-person, second-person, third-person), 
degree of emotional involvement,
focus on self vs.\@ others,
cognitive vs.\@ emotional distance, etc.
Past psychology and mental health research has shown a strong relationship between higher usage of first-person singular pronouns (e.g., \textit{I}) and depression, as well as anxiety; whereas 
first-person plural (e.g., \textit{we}) is less frequent  in people with depression and anxiety \cite{campbell2003secret}.
With WorryWords and TUSC we now explore the degree to which people use anxiety- and calmness-associated words in tweets that include various pronouns.\footnote{We do not include an analysis for the pronoun `us' because the ABCDE datasate preprocessing converted all strings to lower case. Thus the distinction between the country `US' and the pronoun `us' is lost.} 

Figure \ref{fig:anx-pron} (a) shows the percentage of tweets pertaining to each of the pronouns. Observe that the most frequent pronoun in English tweets is \textit{I} ($\sim$60\%),  followed by \textit{you} ($\sim$31\%).
Figure \ref{fig:anx-pron} (b) shows the average anxiety scores for tweets that include each of the pronouns. 
The aggregate anxiety score for all tweets and the aggregate anxiety score for all pronoun tweets are shown as the two dashed lines at $-$15.13 and $-$17.77, respectively; it is worth noting that tweets with pronouns have lower anxiety score than tweets in general (statistically significant). This may be due to a number of reasons such as: the bias to come across as happy when talking about oneself and to not freely express anxiety in tweets that refer to a specific person.

Observe that tweets with \textit{her, we, you,} and \textit{she} have average anxiety scores lower than the average for pronouns; whereas tweets with \textit{he, they, me,} and \textit{I} have higher anxiety scores.
The pronoun \textit{they} has roughly the same anxiety score as that of all tweets, whereas the pronoun \textit{he} has anxiety score slightly higher than the average for all tweets. These results indicate that third-person mentions include more anxiety (fewer calmness) words than first- and second-person pronouns, with the exception of pronouns referring to females (\textit{she, her}).
We also note that subject pronouns (\textit{I, he, she, they}) have higher anxiety scores than their corresponding object pronouns (\textit{me, him, her, them}). 

\section{Conclusion}

We showed that levels of anxiety on social media posts exhibit systematic
patterns of rise and fall during the day and across the week.
We also examined anxiety in past, present, and future tense sentences to show
that anxiety is highest in past tense and lowest in future tense.
Finally, we examined the use of anxiety and calmness words in posts that contain pronouns to show: 
more anxiety in 3rd person pronoun (\textit{he, they}) posts 
and higher anxiety in posts with subject pronouns (\textit{I, he, she, they}) compared to object pronouns (\textit{me, him, her, them}). 
Overall, these trends provide valuable insights on not just when we are anxious, but also how different types of focus (future, past, self, outward, etc.) are related to anxiety. These results also establish baselines that can be compared against when examining specific corpora of interest such as tweets about climate change, vaccines, immigration, hate speech, etc. 
Future work will explore the question of \textit{*when*} we are more anxious in various languages. We hope this line of work will lead to a better understanding of anxiety, both in terms of what is common, as well as how different groups and cultures experience anxiety differently.

\newpage

\section{Limitations}
\label{sec:limitations}

The experiments in Section 6 make use of existing anxiety and social media datasets: WorryWords and TUSC. The WorryWords lexicon was compiled mainly from US and European English speakers. The TUSC dataset has tweets from US and Canadian residents. Thus, conclusions from this work are limited to those demographics. However, many of these conclusions are consistent with results found using traditional approaches in psychology on various demographics. Nonetheless, norms of anxiety expression vary across regions to some extent. This is due to many reasons, but one simple example is that in many parts of the middle-east, the weekend falls on different days than in the rest of the world. Another example is that society may pay greater importance to different things in different parts of the world, leading to more or less anxiety associated with it in different regions.
Regardless, further experiments with datasets pertaining to other locations and demographics are needed to determine which trends are more global and which exhibit substantial local variation. 

\section{Ethics Statement}
\label{sec:ethics}
The use of the two datasets (WorryWords and TUSC) was in accordance with their terms of use and data sheets (provided on their webpages). The resources are freely available for research.

Anxiety expression is complex and nuanced. Additionally, each individual expresses anxiety differently through language, which results in large amounts of variation. 
See \citet{Mohammad23ethicslex} for a discussion of good practises and ethical considerations when using emotion lexicons. 
Our work follows many of those principles, and we discuss below notable ethical considerations when computationally analyzing anxiety. Importantly, we do not make inferences about individuals and their mental state or potential anxiety disorders. We do aggregate level analysis to determine broad trends in general anxiety. Notable ethical considerations that apply include: 

\vspace*{-4mm}
\begin{enumerate}
    \item \textit{Coverage:} WorryWords includes a large number of English words from many sources. Yet, the words included do not equally cover all domains, genres, and people of different locations, socio-economic strata, etc. It likely includes more of the vocabulary of people in the United States with a socio-economic and educational backgrounds that allow for technology access.
    \vspace*{-1mm}
    \item \textit{Word Senses and Dominant Sense Priors:} Words when used in different senses and contexts may be associated with different degrees of anxiety. The entries in WorryWords are indicative of the anxiety associated with the predominant senses of the words. This is usually not problematic because most words have a highly dominant main sense (which occurs much more frequently than the other senses). 
    In specialized domains, some terms might have a different dominant sense than in general usage. Entries in the lexicon for such terms should be appropriately updated or removed. 
    \item  \textit{Socio-Cultural Biases:} The annotations for anxiety capture various human biases. These biases may be systematically different for different socio-cultural groups. WorryWords was annotated by mostly US  English speakers and we applied it to TUSC which includes posts by US residents, but even within the US there are many diverse socio-cultural groups.
    Notable also, crowd annotators on Amazon Mechanical Turk (used to create WorryWords) do not reflect populations at large. In the US for example, they tend to skew towards male, white, and younger people. However, compared to studies that involve just a handful of annotators, crowd annotations benefit from drawing on hundreds and thousands of annotators (such as this work). 
\vspace*{-1mm}
\end{enumerate}


\bibliography{custom}

\appendix

\section{APPENDIX}
\label{appendix-a}

\subsection{Experimental Setup and Other Notes}

\begin{enumerate}

\item If a tweet includes more than one pronoun, then it is included in the bins for each of those pronouns.

\item The pronoun \textit{you} can be used both as the subject and object pronouns, and we speculate that when considering those instances separately, the subject \textit{you} will have a higher anxiety score than the object \textit{you}.

\item Why Twitter (X)? Our long-term goal is to contribute to the broad understanding of anxiety. For that, it is beneficial to examine many different platforms and ranges of time periods. Each of those works is important to make specific claims about those time periods and those modalities, but they also contribute to the broader picture: which trends are stable and which are changing. 
In that context, this work is on US and Canadian Twitter data from 2015 to 2021. 

Additionally, it is a good research practice to tie current results (on specific dataset) with what others have found using different methods, other modalities, and on other time ranges. In this paper, we show how our results fit in with past work in Psychology (based on self-report data) and in Biology (based on cortisol levels). 

\item We deliberately ignore posts from recent years due to the marked rise in AI-generated content online since 2022.

\item Note that higher anxiety scores for 8am and lower scores for 12pm could theoretically be either because (1) people are more anxious in the morning, and more relaxed at noon; Or, (2) people who are more anxious tend to post at 8am; whereas  people who are more relaxed tend to post around noon. Past work in psychology shows evidence towards (1) \cite{moberly2008ruminative,takano2013ruminative,cox2022effects,vedhara2003investigation}.

\item Not everyone has the same work schedule. However, in US and Canada (the source of TUSC tweets), the predominant work schedule is holiday on Saturday and Sunday; and work days (9am to 5pm) Monday to Friday.

\end{enumerate}

\subsection{Computational Resources and Carbon Footprint}
An advantage of using simple lexicon-based approaches is the low carbon footprint and computational resources required. All of the experiments described in the paper were conducted on a regular personal laptop.

\end{document}